\documentclass{article}

\usepackage{amsmath,amssymb}
\usepackage{microtype}
\usepackage{graphicx}
\usepackage{booktabs} 
\usepackage{dsfont}
\usepackage{diagbox}
\usepackage{multirow}
\usepackage{booktabs}
\usepackage{subcaption}
\usepackage[switch]{lineno}
\usepackage{caption}
\usepackage{mathtools}
\usepackage{algorithm}
\usepackage{algorithmic}
\usepackage{hyperref}
\usepackage{pifont}
\usepackage{amssymb}

\usepackage{hyperref}

\renewcommand{\algorithmicrequire}{ \textbf{Input:}} 
\renewcommand{\algorithmicensure}{ \textbf{Output:}} 
\usepackage{mathtools}

\DeclarePairedDelimiter\norm{\lVert}{\rVert}%

\usepackage[accepted]{icml2021}

\icmltitlerunning{Backdoor Scanning for Deep Neural Networks through K-Arm Optimization}

\begin{document}

\twocolumn[
\icmltitle{Backdoor Scanning for Deep Neural Networks through K-Arm Optimization}



\icmlsetsymbol{equal}{*}

\begin{icmlauthorlist}
\icmlauthor{Guangyu Shen}{equal,to}
\icmlauthor{Yingqi Liu}{equal,to}
\icmlauthor{Guanhong Tao}{to}
\icmlauthor{Shengwei An}{to}
\icmlauthor{Qiuling Xu}{to}
\icmlauthor{Siyuan Cheng}{to}
\icmlauthor{Shiqing Ma}{goo}
\icmlauthor{Xiangyu Zhang}{to}
\end{icmlauthorlist}

\icmlaffiliation{to}{Department of Computer Science, Purdue University, West Lafayette, IN, USA}
\icmlaffiliation{goo}{Department of Computer Science, Rutgers University, Piscataway, NJ, USA}

\icmlcorrespondingauthor{Guangyu Shen}{shen447@purdue.edu}

\icmlkeywords{Machine Learning, ICML}

\vskip 0.3in
]

\newcommand{\todoc}[2]{{\textcolor{#1}{\textbf{#2}}}}
\newcommand{\todored}[1]{{\todoc{red}{\textbf{[#1]}}}}
\newcommand{\todogreen}[1]{\todoc{green}{\textbf{[#1]}}}
\newcommand{\todoblue}[1]{\todoc{blue}{\textbf{[#1]}}}
\newcommand{\todoorange}[1]{\todoc{orange}{\textbf{[#1]}}}
\newcommand{\todobrown}[1]{\todoc{brown}{\textbf{[#1]}}}
\newcommand{\todogray}[1]{\todoc{gray}{\textbf{[#1]}}}
\newcommand{\todopink}[1]{\todoc{pink}{\textbf{[#1]}}}
\newcommand{\todopurple}[1]{\todoc{purple}{\textbf{[#1]}}}

\newcommand{\xz}[1]{\todored{XZ: #1}}
\newcommand{\siyuan}[1]{\todoblue{Siyuan: #1}}
\newcommand{\yl}[1]{\todogreen{Yingqi: #1}}
\newcommand{\gy}[1]{\todogreen{gy: #1}}

\newcommand{\cmark}{\ding{51}}%
\newcommand{\xmark}{\ding{55}}%




\printAffiliationsAndNotice{\icmlEqualContribution} 

\begin{abstract}
Back-door attack 
poses a severe threat to deep learning systems. 
It injects hidden malicious behaviors to a model such that any input stamped with a special pattern can trigger such behaviors.
Detecting back-door 
is hence of pressing need.
Many existing defense techniques use optimization to generate the smallest input pattern that forces the model to misclassify a set of benign inputs injected with the pattern to a target label.
However, the complexity is quadratic to the number of class labels
such that they can hardly handle models with many classes. 
Inspired by  Multi-Arm Bandit in Reinforcement Learning, we propose a K-Arm optimization method for backdoor detection. By iteratively and stochastically selecting the most promising labels for optimization with the guidance of an objective function, 
we substantially reduce the complexity, allowing to handle models with many classes. Moreover, by iteratively refining the selection of labels to optimize, it 
substantially mitigates the uncertainty in choosing
the right labels, improving 
detection accuracy.
At the time of submission, the evaluation of our method on over 4000 models in the IARPA TrojAI competition from round 1 to the latest round 4  achieves top performance 
on the leaderboard.
Our technique also supersedes five state-of-the-art techniques in terms of accuracy and the scanning time needed. The code of our work is available at \url{https://github.com/PurduePAML/K-ARM_Backdoor_Optimization}
\end{abstract}

\section{Introduction}
The semantics of a deep neural network is determined by model parameters that are not interpretable. 
Trojan (back-door) attack exploits the uninterpretability and injects malicious hidden behaviors to neural networks. 
To activate back-door behavior, 
the attacker stamps a {\em trigger} to a benign input and passes the stamped input to the trojaned model, which then misclassifies the input to the {\em target label}.
When benign inputs are provided, the trojaned model has comparable accuracy as the original one.
The feasibility of trojan attack has been demonstrated by many existing works. For example, data poisoning~\cite{gu2017badnets} directly uses stamped inputs in training to inject back-door. Neuron hijacking~\cite{liu2018trojaning} compromises a small number of selected neurons by changing their associated weight values through input reverse engineering and retraining. Clean-label attack~\cite{shafahi2018poison} injects malicious features to the target class samples instead of victim class samples, and hence is more stealthy. 
More discussion can be found in the related work section.

Realizing the prominent threat, researchers have developed a number of defense techniques that range from detecting malicious (stamped) inputs at runtime~\cite{ma2019nic} to offline model scanning for possible back-doors~\cite{liu2019abs, wang2019neural, kolouri2020universal}. The former is an on-the-fly technique and requires the presence of malicious inputs. The latter determines if a given model contains any backdoor. It usually assumes a small set of benign inputs for all 
the classes of the model but not any malicious inputs.
Existing scanners usually consider two types of backdoors.
The first is {\em universal backdoor} that causes misclassification (to the target label) for benign samples from any class when they are stamped with the trigger. 
The second is {\em label-specific backdoor} that only causes misclassification of benign samples from a specific {\em victim class} to the target label, when they are stamped with the trigger.
{\em Neural Cleanse} (NC)~\cite{wang2019neural} uses optimization to derive a trigger for each class and observes if there is any trigger that is exceptionally small and hence likely injected instead of naturally occurring feature. {\em Artificial Brain Stimulation} (ABS)~\cite{liu2019abs} systematically intercepts and changes internal neuron activation values on benign inputs, and then observes if consistent misclassification can be induced. If so, the corresponding neurons are considered compromised and used to reverse engineer a trigger. More existing techniques are discussed in the related work section.
Although the effectiveness of existing solutions has been demonstrated, they have various limitations.
In particular, since the target label is unknown beforehand, scanners such as NC try to scan all labels. If the backdoor is  label-specific,
the computation complexity is quadratic.
As such they can hardly handle models with many classes.
For example, 
NC cannot finish scanning a TrojAI round 2 model with 23 classes within 15 hours.
Techniques like ABS leverages additional analysis to pre-select a set of labels/neurons to optimize.
However, their effectiveness hinges on 
the correctness of pre-selection.


\begin{figure}[t]
    \centering

    \begin{subfigure}[t]{0.47\linewidth}
        \centering
        \includegraphics[width=\linewidth]{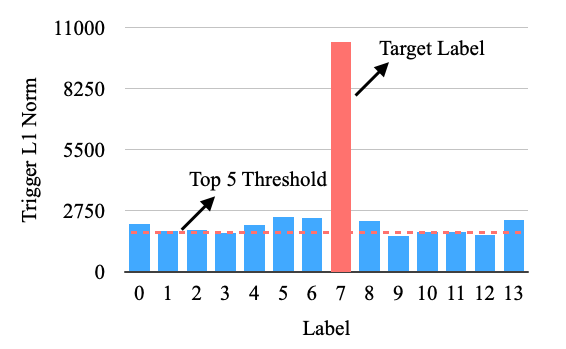}
        \caption{Pre-selection}
        \label{fig:fig2}
    \end{subfigure}
    \begin{subfigure}[t]{0.47\linewidth}
        \centering
        \includegraphics[width=\linewidth]{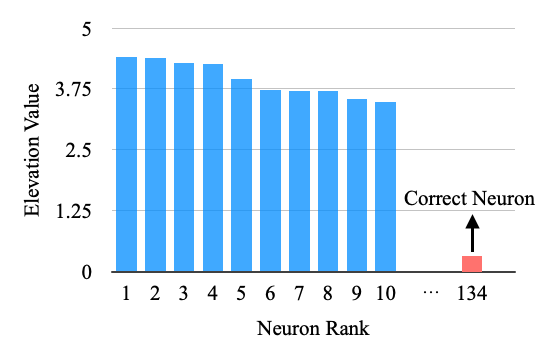}
        \vspace{-15pt}
        \caption{ABS}
        \vspace{-10pt}
        \label{fig:fig3}
    \end{subfigure}
    \vspace{-5pt}
    \caption{\textbf{Motivation cases:} 
    (a) illustrates pre-selection fails to identify backdoor in Model \#56 in TrojAI round 2; (b) shows that ABS fails to identify backdoor in  Model \#13 in TrojAI round 1. 
    }
    \vspace{-15pt}
\end{figure}

\begin{figure*}[t]
    \centering
    \begin{subfigure}[t]{0.23\linewidth}
        \centering
        \includegraphics[width=0.75\linewidth,height=70pt]{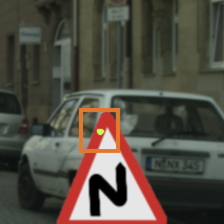}
        \caption{(R4 model \#556) victim class \#13 input + trigger}
        \label{fig:troj_vic}
    \end{subfigure}%
    ~~~
    \begin{subfigure}[t]{0.23\linewidth}
        \centering
        \includegraphics[width=0.75\linewidth,height=70pt]{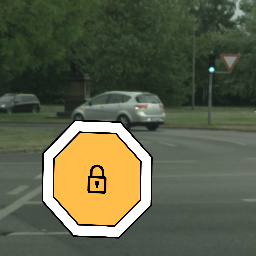}
        \caption{(R4 model \#556) target class \#1 input}
        \label{fig:troj_tar}
    \end{subfigure}
    ~~~
    \begin{subfigure}[t]{0.23\linewidth}
        \centering
        \includegraphics[width=0.75\linewidth,height=70pt]{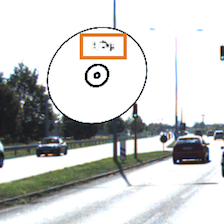}
        \caption{(R4 model \#262) class \#4 input + generated natural feature}
        \label{fig:ben_vic}
    \end{subfigure}
    ~~~
    \begin{subfigure}[t]{0.23\linewidth}
        \centering
        \includegraphics[width=0.75\linewidth,height=70pt]{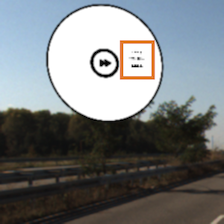}
        \caption{(R4 model \#262) class \#20 input + generated natural feature}
        \label{fig:ben_tar}
    \end{subfigure}
    \vspace{-10pt}
    \caption{\textbf{Motivation cases:} 
    (a) illustrates a victim class \#13   input of a round 4 (R4) {\em trojaned} model stamped with trigger generated by K-Arm, yielding the classification result of label \#1; (b) shows a target class \#1 input for the same model; (c) shows a class \#4 input of a {\em clean} R4 model stamped with natural features generated by K-Arm, yielding label 20; (d) shows a class \#20 input stamped with generated natural features for the same model in (c), yielding label 4.
    }
    \vspace{-15pt}
\end{figure*}

We propose a new back-door scanning method that can handle models with many classes and has better effectiveness and efficiency than existing solutions.
Inspired by {\em K-Arm bandit}~\cite{auer2002finite} in Reinforcement Learning that optimizes decision making with a large number of possible options, we propose a K-Arm backdoor scanner. Instead of optimizing for all the labels one-by-one, the process is divided to many rounds and in each round, our algorithm selects one to optimize for a small number of epochs. 
The selection is stochastic, guided by an objective function. 
The function measures the past progress of a candidate label, e.g., how fast a small trigger can be generated to misclassify stamped inputs to the label, as a trigger is generally easy to optimize if the label is trojaned, 
and how small the trigger is. The stochastic nature of the method ensures that even if the true target label is not selected for the current round, it still has a good chance to be selected later. To our knowledge, we are the first to bring reinforcement learning (K-Arm Bandit) into the neural backdoor detection domain and substantially improve the scanner's efficiency and capability. 
    Natural features sometimes behave
    similarly to backdoors. To distinguish the two, we develop a symmetric optimization algorithm that piggy-backs on the K-Arm backbone. It leverages
    the following observation: while it is easy to optimize a trigger that flips victim label to target label, the inverse (i.e., optimize a trigger that flips target label to victim label) is difficult; natural features, however, do not have this property. 
    
    We evaluate our prototype on 4000 models from IARPA TrojAI round 1 to the latest round 4 competitions, and a few complex models on ImageNet. 
    Our technique achieved top performance on the TrojAI leaderboard and reached the round targets on the TrojAI test server for all rounds.
    It is substantially more effective than the state-of-the-art techniques NC, ABS, and ULP~\cite{kolouri2020universal}  by having 31\%, 20\%, and 27\% better accuracy, respectively. In addition, 
    its scanning time is a few times to orders of magnitude smaller than 
    other optimization based methods, especially in scanning label-specific backdoors.

\section{Related Work}

Besides the ones mentioned in the introduction, we further briefly discuss additional related work and our threat model. 

\noindent
\textbf{Trojan Attack.} Several data-poisoning like attacks \cite{gu2017badnets,liu2018trojaning} utilize  patch/watermark triggers. Clean-label attacks \cite{shafahi2018poison,saha2020hidden,turner2019label,zhao2020clean,zhu2019transferable} inject back-door without changing data label. \citet{salem2020dynamic,nguyen2020inputaware} leveraged generative models to construct dynamic triggers with random patterns and locations for specific samples. 
Composite attack~\cite{lin2020composite} uses natural features from multiple labels as triggers.  Bit flipping ~\cite{Rakin_2019_ICCV,rakin2020tbt} injects malicious behaviors by flipping bits in model weights. Trojan attacks have been developed for transfer learning~\cite{rezaei2019target,wang2018great,yao2019latent}, federated learning~\cite{bagdasaryan2020backdoor,xie2019dba,wang2020attack} and NLP tasks ~\cite{chen2020badnl,sun2020natural}.   

\noindent
\textbf{Existing Detection.} 
ULP~\cite{kolouri2020universal} trains a classifier to determine if a model is trojaned. It leverages a large pool of benign and trojaned models to learn a set of universal input patterns that can lead to different logits for benign and trojaned models. The classifier is then trained on these logits. 
Similar to ULP \cite{kolouri2020universal}, researchers in \cite{huang2020one} proposed one-pixel signature. They trained a classifier to predict the model's benignity based on their one-pixel signature. \citet{qiao2019defending} proposed to generate trigger distribution. \citet{zhang2020cassandra,wang2020practical} leveraged the differences of adversarial examples for benign and trojaned models to detect backdoors.   TABOR \cite{guo2019tabor} used
explainable AI techniques to scan backdoors. \citet{xu2019detecting} detected backdoors using Meta Neural Analysis. \citet{liu2018fine} combined pruning and fine-tuning to weaken or even eliminate backdoors. \citet{wang2020certifying} certified model robustness against backdoor via randomized smoothing. 
\citet{chan2019poison,gao2019strip,chen2018detecting,chou2020sentinet,du2019robust,liu2017neural,ma2019nic} aimed to detect if a provided input contains trigger. Comprehensive surveys of backdoor learning can be found at~\cite{li2020deep,li2020backdoor} 

\noindent
\textbf{Multi-Arm Bandit.} Multi-Arm Bandit (MAB) describes the dilemma of making a sequence of decisions to maximize reward, which has an unknown distribution. It has been thoroughly studied in \cite{auer2002finite}. Many solutions are proposed to tackle this problem, such as Upper Confidence Bound (UCB) \cite{auer2002using}, $\epsilon$-greedy \cite{watkins1989learning}, etc. 
MAB is a general idea with many applications, Our design is inspired by MAB and unique for backdoor detection.

\textbf{Threat Model} We consider a standard setting in the backdoor scanning. Given a model and a small set of clean images without trigger information for each class (less than 20), the defender is required to identify whether the model is trojaned or not. In this paper, we mainly discuss the backdoor with limited size on the propose of stealthiness, such as patch triggers~\cite{liu2018trojaning} or small perturbations~\cite{saha2020hidden}. The injected backdoor can be static~\cite{gu2017badnets}, input aware dynamic~\cite{nguyen2020inputaware}, label-specific or global. Large triggers such as the composite attack~\cite{lin2020composite} and filter triggers~\cite{liu2019abs,cheng2020deep} are out of the scope. We will leave it to the future work.

\section{Motivation}
\label{sec:motivation}
In this section, we discuss the limitations of existing optimization based backdoor scanners and motivate ours.

\noindent
 \textbf{NC~\cite{wang2019neural} cannot handle models with many classes.} Assume a model has $N$ classes. 
 Since the target label is unknown, 
 to detect universal backdoors, NC considers each of the $N$ labels could be the target label and optimizes a trigger that flips benign samples from any class to the label.
 To detect label specific backdoors, it considers each pair of labels could be the victim and target labels, and optimizes a trigger to flip only samples of the victim class to the target label.
 It then checks if there is an exceptionally small trigger (among all those generated). If so, the model is considered having a backdoor.
 The computation complexity is hence 
 $\mathcal{O}(N)$ for universal backdoors and $\mathcal{O}(N^2)$ for label specific backdoors.
 Our experiment (in Section~\ref{experiment}) shows that to scan a model on ImageNet with a universal backdoor, NC needs more than 55 hours. It certainly 
 cannot handle label-specific backdoors
 on such models.

 \noindent
 \textbf{Pre-selection may miss the correct label(s).}
To address the above limitation,
a pre-selection strategy was proposed in~\cite{wang2019neural} to select a small subset of labels to proceed after 10 steps of optimization. Specifically, it selects the $m$ smallest triggers to continue.
 However, its effectiveness hinges on the correctness of pre-selection, which is difficult to achieve due to the uncertainty in optimization.
 Fig.\ref{fig:fig2} illustrates how 
 pre-selection fails on a TrojAI round 2 model (with a universal backdoor).
 Due to the small time budget allowed for scanning a TrojAI model (600s in round 2), top 5 labels are pre-selected out of 14.
 Observe that the trigger size of the target label is still much larger than most of the other labels after 10 steps and precluded. The situation is
 aggravated when the number of classes is large and backdoors are label-specific.
 In fact, our results show that pre-selection can only achieve 58\%
 accuracy on average in TrojAI rounds 1 to 4 training sets.

\noindent
{\bf ABS may select the wrong neurons in stimulation analysis.}
ABS~\cite{liu2019abs} avoids optimizing for individual labels/label-pairs. It systematically enlarges internal neuron activation values for benign inputs and observes if consistent misclassification (to a certain label) can be achieved. If so, the neurons are considered potentially compromised by trojaning. It then uses optimization to generate a trigger by maximizing the activation values of these neurons.
A model is considered trojaned if the
generated trigger can cause the intended misclassification. 
It works for both universal and label-specific backdoors.
Its effectiveness hinges on correctly identifying the compromised neurons, which has inherent uncertainty as well. 
Fig.~\ref{fig:fig3} shows that for a trojaned model $\#13$ in TrojAI round 1, the top 10 neurons that have the largest elevation for the target label logits when stimulated (and hence cause misclassification to the target label) do not include the truly compromised neuron, which is ranked 134 by the stimulation analysis.  
As such, trigger generation based on the top 10 neurons fails to derive the real trigger. In our experiment,
ABS can only achieve 69\% detection accuracy on average for TrojAI rounds 1 to 4. 

\noindent
{\bf Existing scanners cannot distinguish triggers from natural features.}
Natural features can induce misclassification in a way similar to backdoor triggers. For example, stamping a dog nose to cat images may induce misclassification to dog. As such, optimization based trigger generation like NC and ABS may generate natural features as triggers.
Distinguishing the two is important as 
misclassification caused by natural features is inevitable and a model should not
be blamed for their presence; 
and correctly separating natural features from injected triggers allows model end users to employ proper counter measures.
Many TrojAI models have natural features that behave like triggers. 
Fig.~\ref{fig:ben_vic} presents a benign TrojAI model \#262 in round 4, with a class \#4 input stamped with the natural features generated by K-Arm (i.e., the pixel pattern inside the red box).
It causes the model to misclassify 
to label 20 (shown in (d)).
The inputs to TrojAI models are traffic-sign like foreground objects (e.g., the triangle in Fig.~\ref{fig:troj_vic} and the octagon in Fig.~\ref{fig:troj_tar}) with randomly chosen street-view background. More information can be found in Appendix.D.
Observe classes \#4 and \#20 are similar, and the generated features in (c) resemble the central symbol of class \#20, which explains the misclassification.
Both NC and ABS consider the natural features as a trigger and report the model as trojaned.

\noindent
{\bf Our Method.}
From the above discussion, we can observe that {\em a key challenge lies in the inherent uncertainty in selecting the appropriate label (in NC) or neuron(s) (in ABS) to perform optimization}. An exhaustive method like NC without selection is not effective for complex models while pre-selection and ABS making deterministic choices may fail to select the right one.
The overarching idea of our method is to formulate the whole procedure as a stochastic process in which we continue to make selection at each round. Here and in the rest of the paper, an optimization round does not mean an optimization epoch in the traditional sense but rather finding a smaller trigger (that can cause misclassification).  In particular, a selected label/label-pair/neuron that continues to perform well over time (i.e., whose trigger has been easy to optimize) will have a high probability to be selected in the new round. 
A label/label-pair/neuron that does not get selected in one round has a probability to be selected in the future.
The goal is to allow the true positive to eventually stand out.

Specifically, we start with a {\em warm-up} phase in which we optimize each label (to generate trigger) for a very small number of rounds (2 in this paper).
We retain a history of trigger size variation for each label. 
Then we start the {\em selective optimization}. At each round of selective optimization, we select the label that has the best performance over-time. We use an objective function to measure the performance. For the moment, readers can intuitively consider that we utilize the derivative of trigger size (i.e., how fast the
trigger size changes). 
Note that for a clean label, although the optimization may produce a small trigger at the beginning, it cannot achieve substantial size reduction over time. Therefore, its performance degrades and tends to be replaced. In contrast, although the target label may not perform well at the beginning and hence not be selected, it is eventually selected when the other optimizations get stuck.

\begin{figure}[t]
\centering
\includegraphics[width=\linewidth]{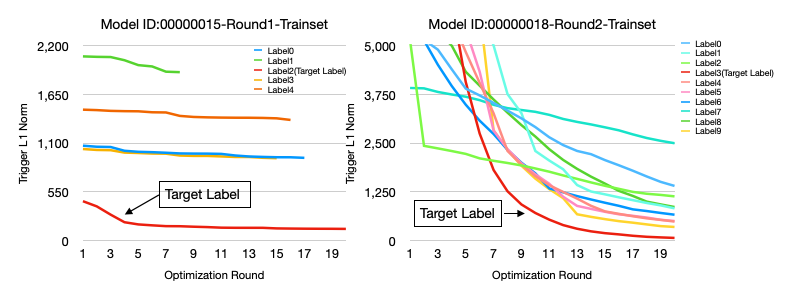}
\vspace{-25pt}
\caption{Trigger size variations over optimization rounds 
}
\vspace{-15pt}
\label{fig:fig4}
\end{figure}

Fig.~\ref{fig:fig4} shows the trigger size variations of all labels over multiple rounds of optimization for two models from TrojAI.
Observe that after the first round, the target label has the smallest trigger for model $\#15$ and hence pre-selection handles it correctly. In contrast for model $\#18$, the target label's trigger is very large and precluded (by pre-selection) from further optimization. Observe that it remains larger than many others till round 5. However, with our method, it eventually stands out and exposes the backdoor. 

The algorithm also seamlessly facilitates separation of natural features and backdoor triggers. 
Specifically, when two benign classes $A$ and $B$ are similar (e.g., cat and dog), small natural features (of $A$) can be identified to flip $B$
samples to $A$ when they are stamped with the features, just like a trigger. Observe 
that since the two classes are similar, small natural
features can be easily identified to flip $A$ to $B$ as well.
For example in Fig~\ref{fig:ben_tar}, 
the generated trigger to flip class \#20 to \#4 has a similar small size as that in Fig.~\ref{fig:ben_vic}.
In contrast, such symmetry is unlikely for real backdoors as generating a trigger to flip the target label to the victim label tends to be difficult.
For example, Fig.~\ref{fig:troj_vic}
 shows a trigger (the pixel in the red box) for model \#556 in round 4 that has a label-specific backdoor from class \#13 to class \#1.  It is sufficient to flip all class \#13 inputs to class \#1 (i.e., Fig.~\ref{fig:troj_tar}). However, due to the differences of the two classes, flipping class \#1 inputs to class \#13 is much more difficult.
Hence, we extend the algorithm such that when it decides to optimize for a victim-target label pair, it also sufficiently optimizes along the opposite direction to check symmetry. 

\section{Design}

\begin{figure}[t]
\centering
\includegraphics[width=\linewidth]{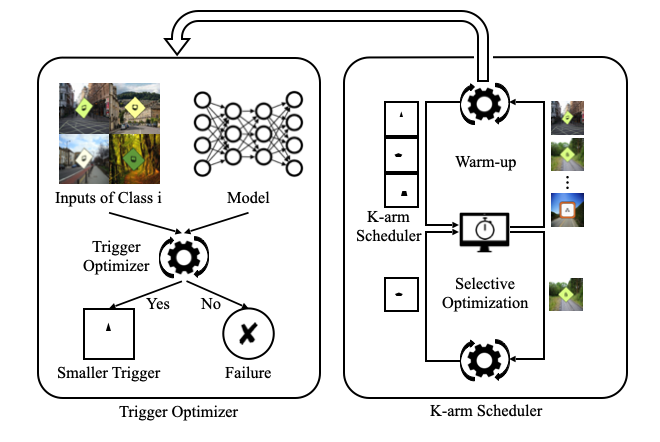}
\vspace{-20pt}
\caption{K-arm optimization workflow}
\vspace{-15pt}
\label{fig:fig5}
\end{figure}

Fig.~\ref{fig:fig5} presents the overview of our technique. On the left is the 
{\em trigger optimizer} (Section~\ref{sec:optimizer}) that performs one round of trigger optimization at a time. In each round the optimizer generates a smaller trigger (than before) that causes a given set of benign samples to be misclassified to a target label, or returns failure when such a trigger cannot be found within a fixed number of epochs. 
On the right is the {\em K-Arm scheduler} (Section~\ref{sec:scheduler}) that decides which {\em arm} should be optimized next.
Assume a model has $N$ classes.
To identify universal backdoor,
we create $N$ (optimization) arms, each
having one of the $N$ labels as the target label and aiming to generate a trigger to flip benign samples from the remaining $N-1$ classes 
to the target label. 
To identify label-specific backdoor,
we create $N\times (N-1)$ arms (i.e., all the pair-wise combinations), each aiming to flip samples of a victim class to a target label. 
Hence, the scheduler selects from the $K=N+N\times(N-1)$ arms. 
In the diagram,
there are two cycles inside the scheduler representing two optimization phases.
The top cycle denotes the {\em warm-up} phase that optimizes all arms for two rounds. 
The scheduler receives and retains the generated trigger information for later use.
The bottom cycle denotes the later {\em selective optimization} phase, in which one selected arm is optimized in each round. 
The selective optimization terminates when we can get a sufficiently small trigger or the time budget runs out. 
To improve efficiency, the scheduler is facilitated by a prescreening phase to reduce unnecessary arms (Section~\ref{sec:prescreening}).  It also considers symmetry during selection to distinguish nature features from triggers (Section~\ref{sec:symmetry}).   

\subsection{Trigger Optimizer}
\label{sec:optimizer}
In each round, the trigger optimizer optimizes one selected arm, generating a trigger for the target label of the arm. Specifically, 
a trigger $T$ is composed of two 
parts: pattern $P$ and mask $M$ with the former deciding the input values of a trigger and the latter deciding the shape/position of the trigger.  
Given a clean input $x$ and a trigger, the stamped input $\hat{x}$ is defined as follows.
\setlength{\belowdisplayskip}{5pt} \setlength{\belowdisplayshortskip}{0pt}
\setlength{\abovedisplayskip}{5pt} \setlength{\abovedisplayshortskip}{0pt}
\begin{equation}
\small
\hat{x} =  
(1-M)\cdot x + M \cdot P
\label{trigger_def}
\end{equation}
Here, operator $\cdot$ stands for the element-wise production. 
Given an $x$ of dimensions $[C,H,W]$, the dimensions of pattern $P$ and $M$ are identical to $x$'s. The values of $P$ are in the range of $[0,255]$ and the values of $M$ are in the range of $[0,1]$. 
Intuitively, stamping a trigger is by mixing $x$ and $P$ through the mask $M$. 
Given a model $\mathcal F$, a target label $t$, and a set of inputs $X$, the trigger optimization for $t$ is defined as follows. 
\begin{equation}
\label{eq:2}
\small
\begin{aligned}
&\min_{P,M} (\mathcal L(t,\mathcal F ((1-M)\cdot X + M \cdot P)) + \alpha\norm{M}_1 )
, \forall x \in X
\end{aligned}
\end{equation}
For an arm of generating universal trigger, $X$ contains a set of clean inputs from classes other than $t$; for an arm of generating label-specific trigger, $X$ contains a set of clean inputs from the victim class. $\mathcal{L}$ stands for the cross-entropy loss function.
Hyper-parameter $\alpha$ balances the attack success rate and the size of the optimized trigger. The optimizer finishes a round and returns if the current trigger $T$ 
satisfies the following condition.
 \[\small Acc( \hat{X},t) \ge \theta\ \mathop{and}\ \norm{M}_1  < \norm{M_p}_1\]
Intuitively, the attack success rate with the trigger needs to be greater than a threshold $\theta$, which is 0.99 in this paper, meaning samples stamped with the trigger have higher than 99\% chance to be classified to $t$, and the current trigger is smaller than the previous one $M_p$.
The optimizer may return failure for the current round when the budget for the label runs out (which is 10 epochs in this paper). 

\subsection{K-Arm Scheduler}
\label{sec:scheduler}

\noindent
To handle  uncertainty in arm selection, 
we leverage the $\epsilon$-greedy algorithm~\cite{watkins1989learning} to introduce randomness in our selection.
The idea is to draw a random sample from a distribution, which is a uniform distribution from 0 to 1 in this paper. If the sample is larger than a threshold $\epsilon$, we rely on an objective function to make the selection; otherwise, a random arm is selected. 
The procedure of selecting label $L$ is formally defined as follows.
\begin{equation}
\label{eq:3}
\small
\begin{aligned}
L &= \left\{
\begin{aligned}
&\mathop{\arg\max}_{l} \ A(l)  ,\ s > \epsilon\\
&rand(K),\ s < \epsilon\\
\end{aligned}
\right. ,\  \mathop{with}\ s\sim U(0,1) 
\end{aligned}
\end{equation}
The parameter $\epsilon$ decides the level of greediness (or randomness).  
With the $\epsilon$-greedy method, even if the true positive label is not selected in an early round, it still has a chance to be chosen in the following rounds. 
We set $\epsilon = 0.3$ in this paper and will discuss its effect later 
in the section. $A(l)$ is an objective function for the target label $l$ of an arm. 
It is supposed to approximate the likelihood of the label being the true label target. 
We leverage two kinds of information in the approximation: {\em the current trigger size for the label} and {\em the trigger size variation for the label over rounds of optimization}. To simplify discussion, we leave symmetry (to distinguish natural features and triggers) to a later section.
Intuitively, a label with a smaller trigger size is promising, and a label that continuously achieves good trigger size reduction in the past is promising. Let $tm(l)$ be the accumulated time spent on optimizing $l$ (in the past rounds); $M(l)$ the current mask of $l$ such that $\norm{M(l)}_1$, the $L_1$ norm of $M(l)$, describes the trigger size; and $M_1(l)$ the first valid trigger for $l$. The objective function $A(l)$ is hence defined as follows.   
\begin{equation}
\label{eq:4}
\small
\begin{aligned}
A(l) &= 
\frac{\norm{{M}(l)}_1 - \norm{ {M}_1(l)}_1}{tm(l)} + \beta \cdot \frac{1}{\norm{{M}(l)}_1}
\end{aligned}
\end{equation}
Here $\beta$ is a hyper-parameter set to $10^5$. 
In the early rounds, the trigger size reduction rate (i.e., the first term in the above equation) is a stronger indicator of true positive. The equation allows us to put more weight on the reduction rate instead of the trigger size, which tends to be large at the beginning and hence the second term tends to be small.
As the optimization proceeds,
the trigger size reduction rate degrades, even for the true positive label, the second term becomes dominating, allowing the scheduler to prioritize labels with small triggers (to make them smaller).

In the end, we compare the size of the smallest trigger with a threshold $\tau$ to decide whether a model is trojaned or benign. In this paper, we set $\tau = 300$ for all TrojAI models and $\tau = 350$ for ImageNet models. 

\smallskip
\noindent
\textbf{Theoretical Analysis of K-Arm.}
We conduct theoretical analysis to show that  K-Arm is more effective (i.e., having higher accuracy) and more efficient (i.e., lower overhead) than NC and NC+pre-selection. The effectiveness is proved by computing the expected time of finishing trigger generation for the true target label.
Details can be found in Appendix.A.

\subsection{Arm Pre-screening}
\label{sec:prescreening}
According to the theoretical analysis, when the number of arms $K$ is large, the cost is dominated by the warm-up phase that is determined by $K$. A large $K$ is hence undesirable. Recall that for a model with $N$ classes, $K=N+N\times(N-1)$,
which could be large. We hence propose a pre-screening step to filter out arms that are not promising. 

In order to achieve high attack success rate, the attacker often has to stamp many benign samples (of various classes when injecting a universal backdoor) with the trigger and use them in trojan training. 
Note that these stamped samples have their labels set to the target label.
As such, the model learns the correlations between the target label and the benign features belonging to the original labels. 
Consequently, the logits value of the target label tends to be {\em consistently} larger than other labels for benign samples. We leverage this to preclude labels that do not look promising.

Specifically, for universal backdoor scanning, we consider a label promising if its logits value ranks among the top $\gamma\%$ labels in at least $\theta\%$ of all the benign samples (of various labels) that can be leveraged for scanning. Collecting such statistics has much lower cost compared to optimization. We set $\gamma= 25$ and $\theta=65$ in this paper.
For label-specific backdoor scanning, we consider an optimization arm from the victim label $t_s$ to the target label $t_d$ promising if $t_d$'s logits value ranks among the top $\gamma\%$ labels in at least $\theta\%$ of all the available benign samples of label $t_s$. We set $\gamma= 25$ and $\theta=90$ in this paper. Observe that our settings of $\gamma$ and $\theta$ are conservative in order not to exclude the right one. We also empirically study the effect of different settings.

According to our experiments in the next section, the pre-screening can substantially reduce the number of arms to consider. For example, we can effectively reduce the arms of ImageNet from $1000$ to $20$ without sacrificing accuracy in universal backdoor scanning. 

\subsection{Symmetric Optimization to Distinguish Natural Features from Triggers}
\label{sec:symmetry}
Assume a (small) trigger $T$ is generated to flip 
clean samples with label $t_s$ to label $t_d$. 
As discussed in Section~\ref{sec:motivation}, If $T$ does not denote a backdoor but rather natural features, the two classes
are likely close to each other.
As such, the trigger flipping samples of $t_d$ to $t_s$ shall have a similar size
as $T$. If $T$ indeed denotes a backdoor, the trigger flipping $t_d$ to $t_s$ tends to be much larger as it is difficult to cause misclassification along the opposite direction of trojaning. 
Therefore, the scheduling algorithm is enhanced as follows to consider symmetry.
The extension focuses on label-specific optimization as such confusion rarely happens for universal backdoors.

Given a label-specific arm $\langle t_s,t_d\rangle$, i.e., flipping $t_s$ to $t_d$, $M(t_s,t_d)$ and $P(t_s,t_d)$ denote the mask and pattern for the generated trigger, respectively, and
$M(t_d,t_s)$ and $P(t_d,t_s)$ the correspondence along the opposite direction (i.e., flipping $t_d$ to $t_s$).
The objective function is as follows.
\begin{equation}
\tiny
\begin{aligned}
A(t_s,t_d) &= 
\frac{(\norm{{M}(t_s,t_d)}_1 - \norm{ {M}_1(t_s,t_d)}_1) / {tm(t_s,t_d)} + \beta \cdot 1 /   \norm{{M}(t_s,t_d)}_1}{(\norm{{M}(t_d,t_s)}_1 - \norm{ {M}_1(t_d,t_s)}_1) / {tm(t_d,t_s)} + \beta \cdot 1 /   \norm{{M}(t_d,t_s)}_1}
\end{aligned}
\end{equation}
Intuitively, we leverage the ratio of objective functions in Equation (~\ref{eq:4}) in the two directions to estimate the likelihood of  $\langle t_s,t_d \rangle$ being the true victim-target label pair.  When $A(t_s,t_d)$ is large, 
meaning the two directions are asymmetric, the pair is likely the true victim-target pair and selected.



    
\section{Experiments}
\label{experiment}
We compare our method with five state-of-the-art techniques against three different attack methods 
on multiple datsets and show that K-arm optimization can achieve better accuracy with lower time cost.

\subsection{Datasets}
\noindent
\textbf{TrojAI Competition.} TrojAI~\cite{trojai} is a program by IARPA that aims to tackle the back-door detection problem.
In each round of competition, the performers are first given a large set of training models (over 1000) with different structures and different classification tasks.
Roughly half of them are trojaned and their malicious identities are known. A (small) set of benign examples are provided
for each label of each model.
These models may be trojaned with various kinds of backdoors, including universal and label-specific. The triggers could be pixel patterns (e.g., polygons with solid color) and Instagram filters~\cite{liu2019abs}. 
They could be position dependent or independent. Position dependency means that the trigger has to be at a specific relative position with the foreground object in order to cause misclassification. A model may have one or more backdoors.
The complexity of models and backdoors grows from round to round.
{\em Note that our technique does not require training. We hence use these training sets as  regular datasets.} 
IARPA also hosts a test set online that is drawn from the same distribution as the training models. It is unknown which test models are trojaned. 
One can submit his/her solution which will be evaluated remotely on their server. The solution needs to finish scanning all the test models (100, 144, 288, and 288 for rounds 1-4, respectively) within 24 hours for rounds 1-2 and 48 hours for rounds 3-4.
By the time of submission,
round 4 is the latest.
We compare our method with the baselines on
all the models with polygon backdoors, mixed with all the clean models across all four rounds.
We exclude models trojaned with Instagram filters as some baselines do not support them. 
The leaderboard results for our technique including both polygon and filter backdoors will be discussed in Section~\ref{experiment_result}.
The details of datasets can be found in Appendix.B.


\begin{table*}[t]
\vspace{-0.1in}
\caption{TrojAI Training Set Results; ``Sym K-Arm Opt + Pre-Srn'' stands for symmetric K-Arm with pre-screening.}
\label{Comparison_small}
\centering
\scalebox{0.8}{
\footnotesize
\tabcolsep=4.9pt
\begin{tabular}{l|rrrr|rrrr|rrrr|rrrr}
\toprule
 &\multicolumn{4}{c|}{Round1} &\multicolumn{4}{c|}{Round2} &\multicolumn{4}{c|}{Round3} &\multicolumn{4}{c} {Round4}\\
\midrule
Method      &Acc &Loss   &ROC   &Time(s)     &Acc &Loss   &ROC   &Time(s) &Acc &Loss   &ROC   &Time(s) &Acc &Loss   &ROC   &Time(s)\\



\midrule

NC &72\% &0.61    &0.73   &623.9 &-   &-   &-   & $>$ 30000  &-   &-   &-   & $>$ 30000  &-   &-   &-   & $>$ 30000\\


Pre-selection  &71\%  &0.62   &0.72   &507.5        &51\%  &1.16   &0.54   &3708.2    &58\%   &0.81   &0.61   &3482.5     &55\%   &1.09   &0.55   &3210.4\\
ABS  &67\%  &0.67   &0.70    &542.9   &62\%   &0.76   &0.57    &1527.0    &71\%   &0.62  &0.56  &1435.0  &79\%  &0.52  &0.55  &525.0\\

TABOR  &80\%  &0.51    &0.81    &1142.2    &55\%    &1.09    &0.59    & $>$ 32000  &60\%    &0.77    &0.57    & $>$ 30000    &60\%    &0.81    &0.55    & $>$ 35000          \\

DLTND   &85\%    &0.45    &0.86    &1109.6    &60\%    &0.79    &0.62    & $>$ 26000    &65\%    &0.75    &0.61     & $>$ 29000    &65\%    &0.77    &0.64    & $>$ 31000        \\ 

\textbf{K-Arm Opt}  &\textbf{90}\%   &\textbf{0.32}   &\textbf{0.90}    &\textbf{275.5}    &\textbf{76}\%   &\textbf{0.58}   &\textbf{0.77}    &\textbf{1956.5}     &\textbf{79}\%   &\textbf{0.50}  &\textbf{0.80}    &\textbf{1740.3}  &\textbf{82}\%   &\textbf{0.51}  &\textbf{0.81}    &\textbf{1623.5} \\

\textbf{K-Arm Opt + Pre-Srn}  &\textbf{-}   &\textbf{-}   &\textbf{-}    &\textbf{-}    &\textbf{75}\%   &\textbf{0.59}   &\textbf{0.76}    &\textbf{140.8}     &\textbf{79}\%   &\textbf{0.50}  &\textbf{0.80}    &\textbf{166.2}  &\textbf{80}\%   &\textbf{0.53}  &\textbf{0.79}    &\textbf{110.5} \\  

\textbf{Sym K-Arm Opt + Pre-Srn}  &\textbf{-}   &\textbf{-}   &\textbf{-}    &\textbf{-}    &\textbf{89}\%   &\textbf{0.33}   &\textbf{0.89}    &\textbf{340.5}     &\textbf{91}\%   &\textbf{0.31}  &\textbf{0.91}    &\textbf{290.5}  &\textbf{89}\%   &\textbf{0.32}  &\textbf{0.89}    &\textbf{204.4} \\  


\bottomrule
\end{tabular}
}
\vspace{-15pt}
\end{table*}

\noindent
\textbf{ImageNet.} We also use 7 VGG16 models on ImageNet (1000 classes) trojaned by TrojNN~\cite{liu2018trojaning}, a kind of unviersal patch attack, and 6 models on ImageNet poisoned by hidden-trigger  backdoors~\cite{saha2020hidden}, with different structures including VGG16, AlexNet, DenseNet, Inception, ResNet and SqueezeNet. The hidden-trigger backdoors are label-specific. They are mixed with 7 clean ImageNet models.

\textbf{Other datasets.} We also evaluate our method on 4 CIFAR10 and 4 GTSRB models trojaned by Input-Aware Dynamic Attack~\cite{nguyen2020inputaware}. They are mixed with 4 clean models respectively.

\vspace{-1pt}
\subsection{Evaluation Metrics}
We report two accuracy metrics used in TrojAI: {\em cross-entropy loss}~\cite{murphy2012machine} and {\em ROC-AUC} (Area under Receiver Operating Characteristic Curve)~\cite{fawcett2006introduction}.
The former is the lower the better and the latter is the higher the better.
In addition, we also report the plain accuracy, i.e., the percentage of models that are correctly classified.
We also report the average scanning time for each model.
For fair comparison, comparative experiments 
are all done on an identical machine with a single 24GB memory NVIDIA Quadro RTX 6000 GPU (with the lab server configuration).
Leaderboard results (on TrojAI test sets) were run on the IARPA server
with a single 32GB memory NVIDIA V100 GPU. We use Adam~\cite{kingma2014adam} optimizer with learning rate $0.1$, $\beta$ = \{0.5, 0.9\} for all the experiments.

\vspace{-1pt}
\subsection{Baseline Methods}
We compare K-Arm with the following state-of-the-art detection methods: ABS~\cite{liu2019abs}, NC~\cite{wang2019neural}, NC+pre-selection~\cite{wang2019neural} (or Pre-selection for short), ULP~\cite{kolouri2020universal}, TABOR~\cite{guo2019tabor}, DLTND~\cite{wang2020practical}. For the optimization based methods including ABS, NC, Pre-selection, TABOR and DLTND, we use the same batch size for fair comparison. For NC, Pre-selection and our method, we use the same early stop condition to terminate the optimization.  For ABS, we select top10 neuron candidates after the stimulation analysis and perform the trigger reverse engineering. For  Pre-selection, we set the number of optimization epochs as $max(10,s)$ for each label with $s$  the number of epochs when the first valid trigger is found. Recall Pre-selection performs a few rounds of optimizations and then selects a promising subset to finish. 
We select the top 3 among the 5 labels for round 1 models and the top 20\% labels/label-pairs for rounds 2-4.
For the ImageNet models, we follow \cite{wang2019neural} and select the top 100.
For ULP, we train it on 500 TrojAI round 1 models and test it on the 100 test models. We did not run it on later rounds as it cannot handle model structure variations in those rounds. For TABOR and DLTND, we use the implementation provided by the authors.

\begin{table}[t]
\caption{Results on ImageNet Models} 
\label{Comparison_large}
\centering
\scalebox{0.7}{
\footnotesize
\tabcolsep=4pt
\begin{tabular}{l|rrrr|rrrr}
\toprule
 &\multicolumn{4}{c|}{Hidden Trigger Attack} & \multicolumn{4}{c}{TrojanNN}  \\
\midrule
Method      &Acc    &Loss    &ROC   &Time(s)     &Acc  &Loss    &ROC  &Time(s) \\



\midrule
NC  &- &- &- & $>$1m   &71\%  &0.65   &0.82    &221k \\ 
Pre-selection  &54\% &1.02 &0.62 &171k   &64\%    &0.92   &0.74  &43k    \\
ABS  &100\%   &0.11 &1.00   &389k   &100\%  &0.11  &1.00 &4.9k    \\
\textbf{K-Arm} &\textbf{85\%}  &\textbf{0.33} &\textbf{0.93}    &\textbf{86k}   &\textbf{88\%}   &\textbf{0.38}  &\textbf{0.92}   &\textbf{19k}   \\
\textbf{K-Arm+Pre-Srn}   &\textbf{85\%} &\textbf{0.33} &\textbf{0.93}    &\textbf{2k}  &\textbf{100\%} &\textbf{0.11}  &\textbf{1.00}   &\textbf{224}\\

\textbf{Sym K-Arm+Pre-Srn}   &\textbf{100\%} &\textbf{0.09} &\textbf{1.00}    &\textbf{4k}  &\textbf{-} &\textbf{-}  &\textbf{-}   &\textbf{-}\\


\bottomrule
\end{tabular}
}
\vspace{-10pt}
\end{table}

\subsection{Parameter Tuning}
We evaluate the effects of hyper-parameters, including the following: $\beta$ in the objective function, $\theta$, $\gamma$ in the arm pre-screening and $\epsilon$ in the K-Arm Scheduler. The last one is the threshold $\tau$ which decides if a model is trojaned. We randomly select 40 models (20 benign and 20 trojaned) from round 2 to test our method.  In detail, we pick 5 different values $(10^2,10^3,10^4,10^5,10^6)$ for $\beta$.  For $\epsilon$, we select 10 values ranging from $0.1\sim 0.5$. We use 5 different $\tau$ values from $100 \sim 500$, 3 $\theta$ values from $10\sim 30$ and  3 $\gamma$ values from $50 \sim 80$.
The results are in Appendix.C.

\begin{table*}[t]
\caption{TrojAI Leaderboard Results}

\label{leaderboard}
\centering
\scalebox{0.6}{
\footnotesize
\begin{tabular}{l|rrrr|rrrr|rrrr|rrrr}
\toprule

 &\multicolumn{4}{c|}{Round1} &\multicolumn{4}{c|}{Round2} &\multicolumn{4}{c|}{Round3} &\multicolumn{4}{c} {Round4}\\
 
\midrule

Method  &CE Loss    &ROC   &Time(s) &Rank   &CE Loss    &ROC   &Time(s)  &Rank   &CE Loss    &ROC    &Time(s) &Rank   &CE Loss    &ROC &Time(s)  &Rank  \\ 

\midrule


NC   &-    &-    &T/O  &-  &-    &-  &T/O   &- &-    &-  &T/O  &-  &-    &-  &T/O &-\\


ABS  &0.64(+0.34)  &0.70(-0.21)   &523(+233)   &-   &0.76(+0.44)    &0.53(-0.36)  &508(+18)  &- &0.84(+0.55) &0.56(-0.35)   &599(+367)    &-   &0.87(+0.55)   &0.48(-0.42)   &229(+18)  &-\\

ULP     &1.18(+0.88)   &0.59(-0.32) &0.1(-290)  &-  &-  &-  &-   &-  &-   &-  &-   &- &-  &-   &- &- \\

\textbf{K-Arm}   &\textbf{0.30(-0.00)}   &\textbf{0.91(-0.00)}   &\textbf{290(-0)}  &\textbf{1}   &\textbf{0.35(+0.03)}   &\textbf{0.90(+0.01)}    &\textbf{290(-200)} &\textbf{2}  &\textbf{0.29(-0.00)}   &\textbf{0.91(-0.00)}  &\textbf{232(-0)} &\textbf{1}  &\textbf{0.33(+0.01)}   &\textbf{0.90(-0.00)}  &\textbf{201(-10)} &\textbf{2} \\







\bottomrule
\end{tabular}
}
\vspace{-15pt}
\end{table*}

\subsection{Experimental results}
\label{experiment_result}
\textbf{Results for TrojAI Rounds 1-4 Training Sets.}
Table~\ref{Comparison_small} shows the comparison results on
the aforementioned models from TrojAI rounds 1-4 training sets (3231 models in total).
Columns Acc, Loss, ROC, and Time stand for plain accurcy, cross entropy loss, AUC-ROC, and average scanning time (in seconds) per model, respectively. 
Observe that our method achieves the best accuracy and has the lowest scanning time compared to all the baselines. The best K-Arm methods have 4\%, 27\%, 30\%, 25\% better ROC than the best performance by the baselines for the four respective rounds.
They are also 1.8, 10.8, 8.6, 4.8  times faster than the fastest among the baselines for the four respective rounds. 
This strongly supports the better effectiveness and efficiency of K-Arm.
K-Arm has higher accuracy than Pre-selection and ABS because they have to make deterministic selection (about which labels/neurons to optimize) at the beginning which is difficult when the candidate sets are large (e.g., in label-specific backdoor scanning).
K-Arm has higher accuracy than NC  even though NC is exhaustive.
Besides that NC does not consider symmetry and hence cannot distinguish natural features from injected triggers, its exhaustive nature in many cases also hurts performance as it aggressively optimizes for clean labels, generating many natural features with small size that behave like triggers. TABOR and DLTND encounter the same problem and suffer from huge number of false alarms.


The last three rows in Table 1 and 2 present the ablation study for different components of our method.
The vanilla K-Arm can have 79\% accuracy and 1773s on average (from R2 to R4). K-Arm with pre-screening achieves 78\% accuracy and 138s. Symmetric K-Arm with pre-screening gets 89\% and 278s. Note that vanilla NC only has 57\% with 32000s. Observe that arm pre-screening substantially reduces the scanning time (by an order of magnitude) without sacrificing much accuracy; symmetric optimization is critical to improving accuracy, with 13\%, 11\%, and 10\% ROC improvement for rounds 2-4. Without the symmetric optimization, K-Arm would not be able to reach the round targets (i.e., lower than 0.348 Loss).



\smallskip
\noindent
\textbf{Results for ImageNet Models.}
Table~\ref{Comparison_large} shows the results for the ImageNet models.
Columns 2-5 present results on the 6 models with (label-specific) hidden-trigger  backdoors mixed with 7 benign models; columns 6-9 present results on the 7 models with (universal) TrojNN backdoors, mixed with 7 benign models.
For hidden-trigger backdoors, the best K-Arm has 100\% accuracy. NC could not finish due to the large number of victim-target label pairs.
It took two weeks to scan a model.
Both Pre-selection and ABS have much worse accuracy or scanning time.
For TrojNN backdoors, 
The best K-Arm has 100\% accuracy, higher than most baselines.
Although ABS can also achieve 100\% accuracy, it is 20 times slower than the best K-Arm.
NC and Pre-selection have lower accuracy and much longer scanning time due to the large number of classes and natural features that behave like triggers.

\begin{figure}[t]
\centering
\includegraphics[width=0.85\linewidth]{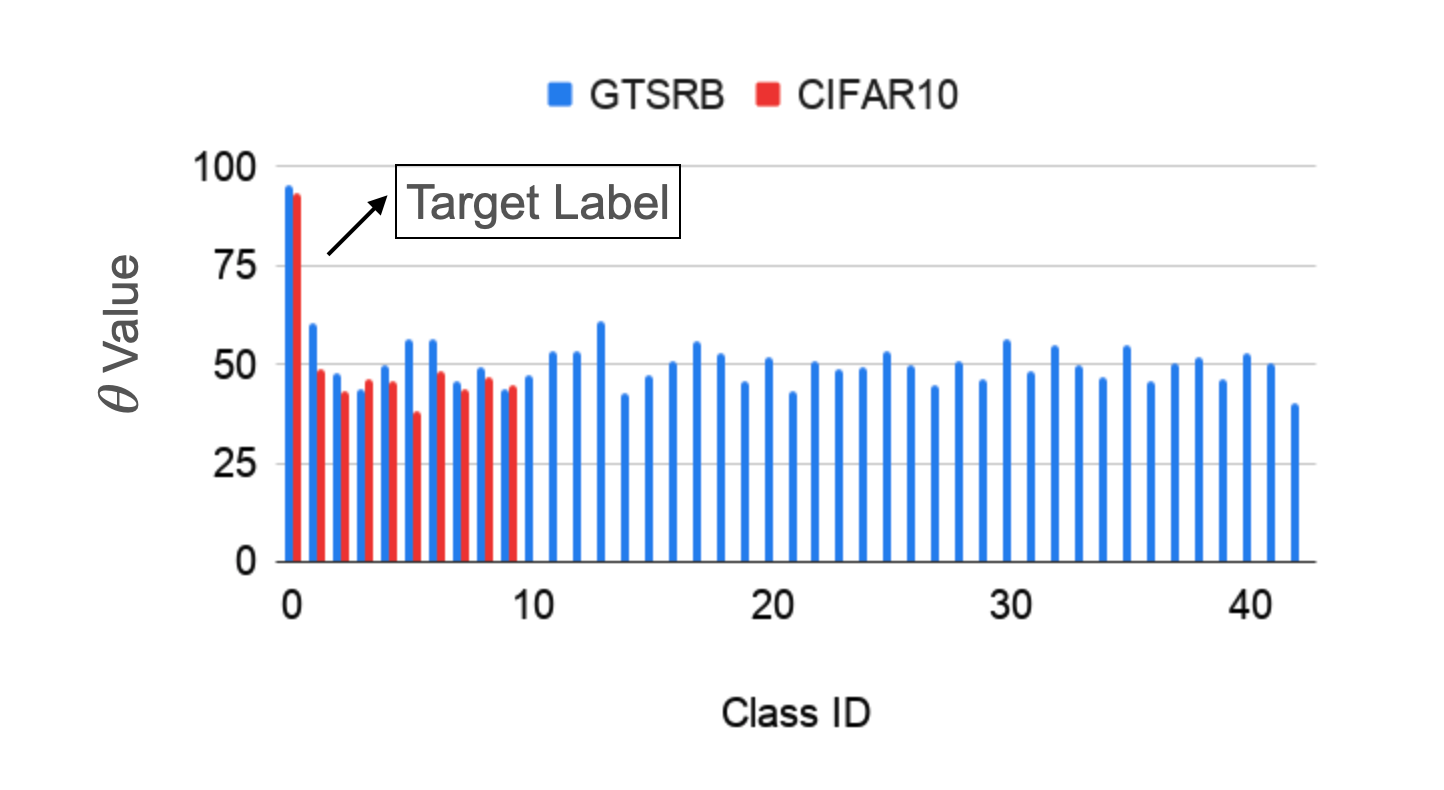}
\vspace{-10pt}
\caption{Results on Input-Aware Dynamic Attack. 
}
\vspace{-20pt}
\label{dynamic}
\end{figure}

\smallskip
\noindent
\textbf{Results for the Dynamic Attack.} Different from static backdoor attacks, dynamic attack can generate input specific triggers. Therefore, the optimized trigger of the target class will not be extremely smaller than others, then bypass the outlier detection. However, our experiment results show that the proposed pre-screening technique can identify the target label preciously for the poisoned models. By setting the bound $\theta = 70$, $\gamma=25$, we can successfully detect all 4 trojan models on CIFAR10 and GTSRB without any false positives. Fig.~\ref{dynamic} shows the $\theta$ values of different classes for a GTSRB and a CIFAR10 poisoned models. The target label is 0 for both models. Observe that the $\theta$ value of the target label is 35\% larger than the largest value of the rest labels, which is a strong indicator for the trojan models. Remind the large $\theta$ reveals that the model learns the target class features as part of the features for other classes due to the poisoning process.

\smallskip
\noindent
{\bf K-Arm Performance on TrojAI Leaderboard.}
K-Arm consistently achieved top results across the four rounds\footnote{\href{url}{https://pages.nist.gov/trojai/}}\footnote{\href{url}{https://pages.nist.gov/trojai/docs/results.html\#previous-leaderboards}}. Table~\ref{leaderboard} shows the K-Arm results for the four rounds, including the loss, ROC, average scanning time, and ranking.
The results include those for all the different types of backdoors (polygon, filter, label-specific, universal, position-dependent, multiple backdoors in a model, etc.).
We also show the difference between K-Arm and the top (if any). For example, in round 2, K-Arm ranked number 2.
Loss 0.35(+0.03) means that K-Arm's loss is 0.35 while the top performer has 0.32 loss; ROC 0.90(+0.01) means that K-Arm has 0.9 ROC while the top performer has 0.89 ROC. Note that the leaderboard ranks solutions by (smaller) loss. 
K-Arm beat the round targets (i.e., lower than 0.348 loss) for 3 out of the 4 rounds. For round 2, although it did not beat the target, its ROC is the highest. 
It ranked number one for 2 out of the 4 rounds. 
In all rounds, K-Arm is faster than ABS. We also train ULP on 500 round 1 training set models and evaluate it on the round 1 test set. However, its accuracy is not high. 
We speculate two reasons: 1) unlike the models in the ULP paper, the classes of TrojAI models are not fixed; 2) the classifier seems to easily overfit on the training data and the triggers in the TrojAI datasets share few common features.
On the other hand,
ULP is not optimization based and hence is extremely fast.

\begin{figure}[t]
\centering
\includegraphics[width=0.85\linewidth]{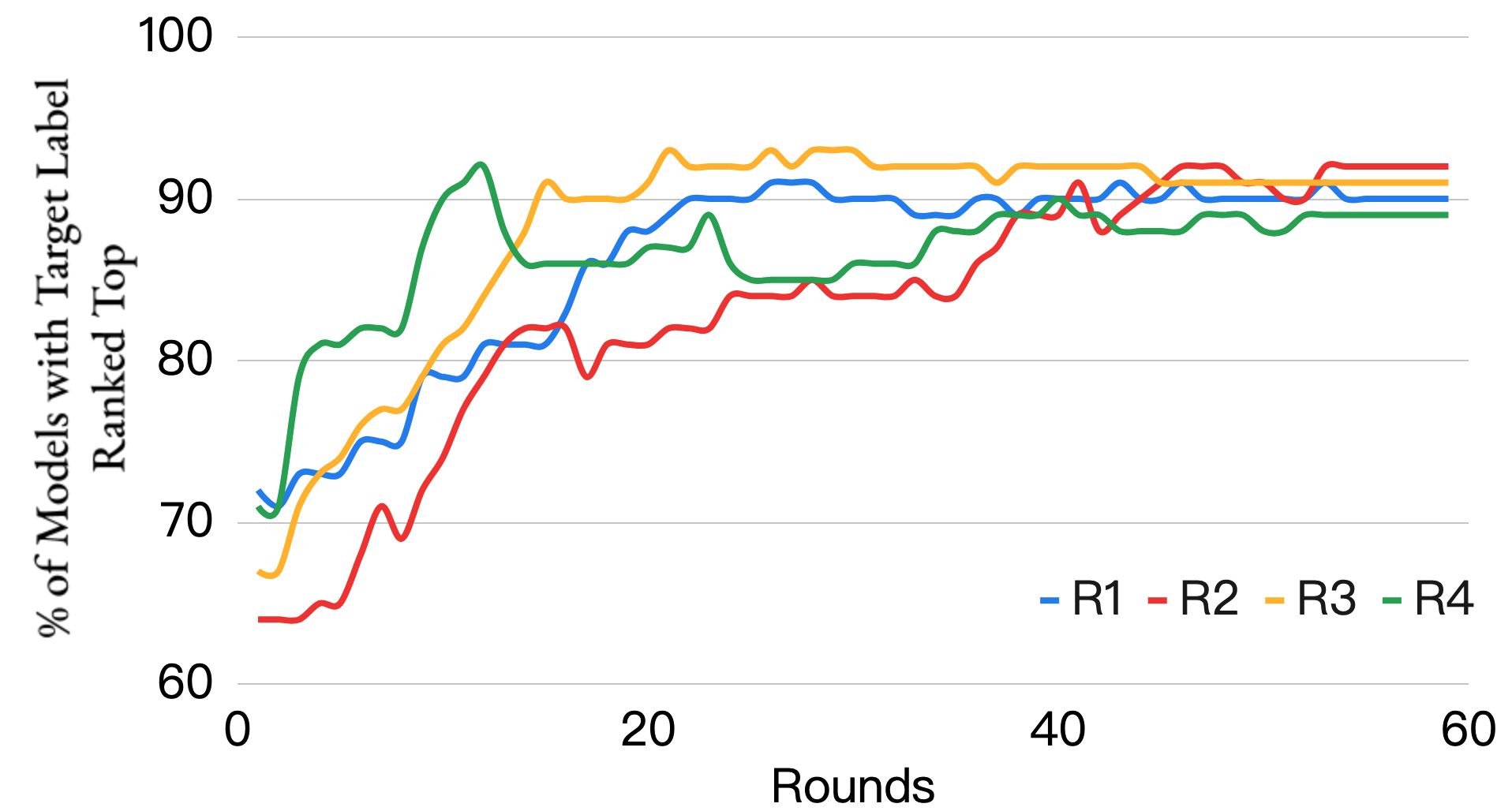}
\vspace{-10pt}
\caption{Trend of Trigger Optimization. 
}
\vspace{-20pt}
\label{trend}
\end{figure}

\smallskip
\noindent
{\bf Trend of Trigger Optimization in K-Arm.}
We randomly sample 100 trojaned models from each training set of TrojAI rounds 1 to 4. We record the ranking of the optimized trigger size of true target label for each model during  optimization. 
Fig.~\ref{trend} shows the percentage of models whose target label trigger size ranks number 1 (i.e., the smallest) for each round. We can see that after warm-up, there are only 60-70\%  models rank top. As such, a simple pre-selection strategy does not work. All the sets converge at around 90\%, indicating that K-Arm allows the true positives to stand out eventually in most cases.
Also observe that the different sets converge at different optimization rounds, indicating that using a universal larger number of warm-up rounds instead of K-Arm will not work. Moreover, 20 rounds of warm-up means hundreds of epochs, which is already not affordable as all arms have to go through warm-up.
At the end, we point out that
there are still around 10\% cases that do not stand out at the end.
We study some of them in the Appendix.D.
We leave the problem to future work.



\noindent
{\bf Adaptive Attack.} We devise an adaptive attack for the arm pre-screening stage.
Our goal is to suppress the target label logits for benign samples of victim classes.
This is done by adding an $\mathcal L_2$ regularization of target label logits value of benign samples.
As such, the optimizer tries to enlarge the distance between the target label and the victim label. 
The strength of the attack is 
controlled by a coefficient.
We use 10 models on CIFAR10 with different coefficient values  and report the model accuracy, attack success rate (ASR), selection accuracy and K-Arm detection accuracy in Table~\ref{adaptive_attack}.
Observe that pre-screening becomes less effective when the attack is stronger. However, the model accuracy and attack success rate degrade as well. It is unclear how to design adaptive attack for the scheduler or optimizer. 
We will leave it to future work.

\begin{table}[t]
\caption{Adaptive Attack} 
\label{adaptive_attack}
\centering
\scalebox{0.85}{
\footnotesize
\tabcolsep=4pt
\begin{tabular}{l|rrrrrr}
\toprule
Coefficient    &0   &1     &10     &100    &1000  \\
\midrule

Model Acc   &80.0\%  &78.1\%     &71.4\%     &65.5\%    &35.4\%\\
ASR         &99.0\%  &99.4\%     &92.9\%   &92.6\%      &0.0\%\\
Selection Acc   &100.0\%  &100.0\%  &80.0\%     &50.0\% &-\\
K-Arm Detection Acc     &100.0\%    &100.0\%  &70.0\% &40.0\%  &-\\




\bottomrule
\end{tabular}
}
\vspace{-10pt}
\end{table}



\section{Conclusion}
Inspired by K-Arm Bandit in Reinforcement Learning, we develop a K-Arm optimization technique for back-door scanning. The technique handles the inherent uncertainty in searching a very large space of model behaviors, using stochastic search guided by an objective function. It shows outstanding performance on models from IARPA TrojAI competitions.
It also outperforms the state-of-the-art techniques that are publicly available.
\section{Acknowledgments}
We thank the anonymous reviewers, Zikang Xiong for their valuable comments.
This research was supported, in part by IARPA TrojAI W911NF-19-S-0012, NSF 1901242 and 1910300, ONR N000141712045, N000141410468 and N000141712947. Any opinions, findings, and conclusions in this paper are those of the authors only and do not necessarily reflect the views of our sponsors.

\nocite{langley00}

\bibliography{example_paper}
\bibliographystyle{icml2021}


\clearpage

\newpage



\end{document}